# Visual Noise from Natural Scene Statistics Reveals Human Scene Category Representations

Michelle R. Greene (1), Abraham P. Botros (1), Diane M. Beck (2) and Li Fei-Fei (1)

(1) Department of Computer Science
Stanford University
mrgreene@stanford.edu; abotros@cs.stanford.edu; feifeili@cs.stanford.edu
(2) University of Illinois Urbana-Champaign
dmbeck@illinois.edu

Author for correspondence:

Michelle R. Greene
Department of Computer Science
Room 240
Stanford University
353 Serra Mall
Stanford, CA 94305

**Abstract**

Our visual perceptions are guided both by the bottom-up information entering our eyes as well as our top-down expectations of what we will see. Although bottom-up visual processing has been extensively studied, comparatively little is known about top-down signals. Here, we describe REVEAL (Representations Envisioned Via Evolutionary ALgorithm), a method for visualizing an observer's internal representation of a complex, real-world scene, allowing us to visualize top-down visual information. REVEAL rests on two innovations for solving this high-dimensional problem: visual noise that samples from natural image statistics, and a computer algorithm that collaborates with human observers to efficiently obtain a solution. In this work, we visualize observers' internal representations of a visual scene category (*street*) using an experiment in which the observer views visual noise and collaborates the algorithm to recreate his internal representation. As no scene information was presented, observers had to use their internal knowledge of the target, matching it with the visual features in the noise. We demonstrate that observers can use this method to re-create a specific photograph, as well as to visualize purely mental images. Critically, we show that the visualized mental images can be used to predict rapid scene detection performance, as each observer had faster and more accurate responses in detecting real-world images that were similar to his template. These results show that it is possible to visualize previously unobservable mental representations of real world stimuli. More broadly, REVEAL provides a general method for objectively examining the content of subjective mental experiences.

**Introduction**

Our visual representations are driven both by the bottom-up input entering our eyes as well as top-down predictions about what we expect to see (1, 2). Although bottom-up visual processing has been extensively studied, comparatively little is known about top-down predictive signals, largely because we have lacked the tools to solicit the content of top-down representations in an unbiased manner. These representations make a bridge between the physical and mental worlds, and understanding the content of these representations will allow us to explore how our experience shapes our expectations.

One can gain insights into a top-down representation by examining the statistical regularities of various natural environments (3–6). Although some of these statistical patterns have been shown to affect behavior (7) and patterns of neural activity (4, 8), this approach is fundamentally limited because the existence of physical differences between images does not necessarily mean that human observers use these differences for perception. Furthermore, this approach does not allow us to explore how an individual's visual experience can subtly shape his representations. Although systematic statistical differences have been observed across different environment types (5), and perception can differ as a function of visual experience (9, 10), we have not been able to link an observer's visual representations to his visual experience in part because we have not been able to visualize these internal representations.

The goal of this work is to create an unbiased visualization of an observer's mental template of a real-world scene. Such visualization would confer numerous advantages to the study of visual recognition. It would enable us to not only to understand the visual features used to perform a task, but also to potentially reveal novel visual coding strategies. An externalized representation would also allow us to study how the use of visual information changes over a range of psychological conditions, such as learning, attention and memory.

A variety of techniques exist to visualize an observer's internal representations, but none scale to the complexity of real-world scenes. In the most popular method, known as "classification images", pixelated Gaussian white noise is used to visualize category boundaries (11, 12). As compelling as these images are, they can only be created with very simple visual classes, and do not scale to the complexity of natural images (13). First, these techniques make strong assumptions that visual information is contained in independent pixels (14), whereas real-world scenes have extended surfaces and objects that violate this assumption. Second, even a relatively small image will have tens of thousands of pixels, requiring hundreds of thousands of trials to adequately sample the space, making this type of experiment is deeply impractical. Last, Gaussian white noise has a very non-natural appearance, and this may bias an observer's classification pattern (15).

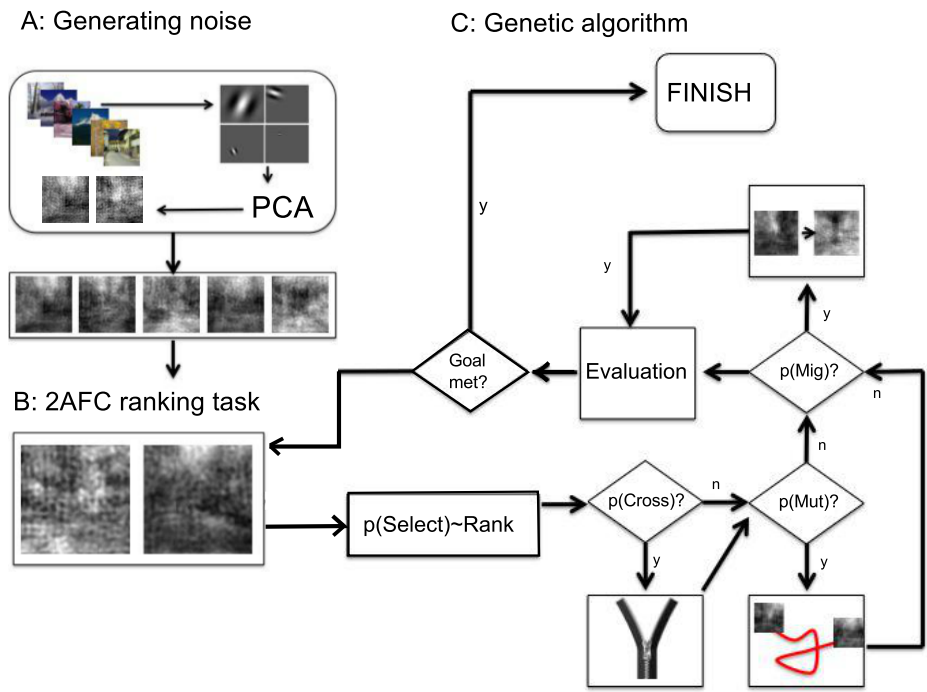

Figure 1: Flowchart of experimental paradigm. (A) Visual noise was created by representing a database of 4200 real-world scenes using a multi-scale Gabor pyramid. Principal components analysis (PCA) was performed on this representation. Noise was generated by randomizing principal component scores for the first 900 components. (B) Human participants completed a 2-alternative forced choice (2AFC) task in which they indicated which of two noise images was most similar to a target. (C) In collaboration with human observers, a genetic algorithm was used to efficiently traverse the noise space and choose images that were more similar to the target. Images were ranked

**from the 2AFC task, and a proto-generation was created by selecting images in proportion to the number of times they were chosen by the observer as being most similar to the target. Next, with a probability of 0.4, two images were randomly selected from the proto-generation for a crossover. The subsequent image contained the odd-numbered principal components of one image and the even-numbered principal components of the other. Next, images were "mutated" with a probability of 0.3. If an image was selected for mutation, its principal components were multiplied by random values scaled to 5% of a standard deviation of the scale of each component. Last, images were selected for migration, or replacement with random images. This probability was 0.6 in the first generation, and decreased by half for each subsequent generation. This new generation of images was then re-ranked by the participant if the reconstruction goal was not yet met.**

In this paper, we introduce REVEAL (Representations Envisioned Via Evolutionary ALgorithm), a novel method for reconstructing and visualizing observers' subjective and unobservable mental representations of real-world scenes. A flowchart of this method can be found in Figure 1. REVEAL rests on two key innovations: a type of visual noise based on the statistical properties of natural scenes, and a genetic algorithm that collaborates with human observers to efficiently sample the high-dimensional noise space. As no actual scene information was ever presented, observers had to match the current visual input with their own internal scene representations in order to perform the task. Therefore, the resulting visual output reflects a visualized reconstruction of the observer's subjective mental image. Importantly, this reconstructed mental image was not only similar to the central tendency of the category, but an individual's template predicted his or her performance in a rapid scene categorization performance. Altogether, REVEAL provides a method for visualizing subjective, previously unobservable mental images of complex, real-world scenes. This allowed us, to predict an observer's perceptions and thus provides us with the critical first step in understanding how an individual's top-down template impacts their perception.

## Results

### *Recreating an image with an ideal observer*

As a proof-of-concept, we simulated the performance of an ideal observer who always chose the optimal image on each trial.  The goal of the simulation was to re-create a specific photograph (see Figure 2A). Critically, this photograph was *not* a part of the scene database used to measure image statistics for creating visual noise (see Methods). This ensures that we can find solutions that are not merely a linear combination of its inputs. This simulation reflects the best-case scenario for our method, and can thus serve as a performance ceiling for comparison with human performance. To ensure that the quality of the end product does not depend on the original state of the image population, we rejected and replaced any randomly generated image that was at or above the 80th similarity percentile from an empirically derived chance distribution (see Methods for details). We ran 100 simulations with the ideal observer. As shown in Figure 2B, the ideal observer achieved a statistically significant (less than 5% of randomly generated images were more similar to target) level of similarity in 6 generations on average (95% CI: 5-7). The ideal observer achieved reconstruction greater than 99.99% of random images in 20 generations (CI: 11-39), and achieved a correlation with the target image of r=0.99 in 1,059 generations (CI: 951-1,186). Therefore, we can fully reconstruct a specific, arbitrary image using our method.

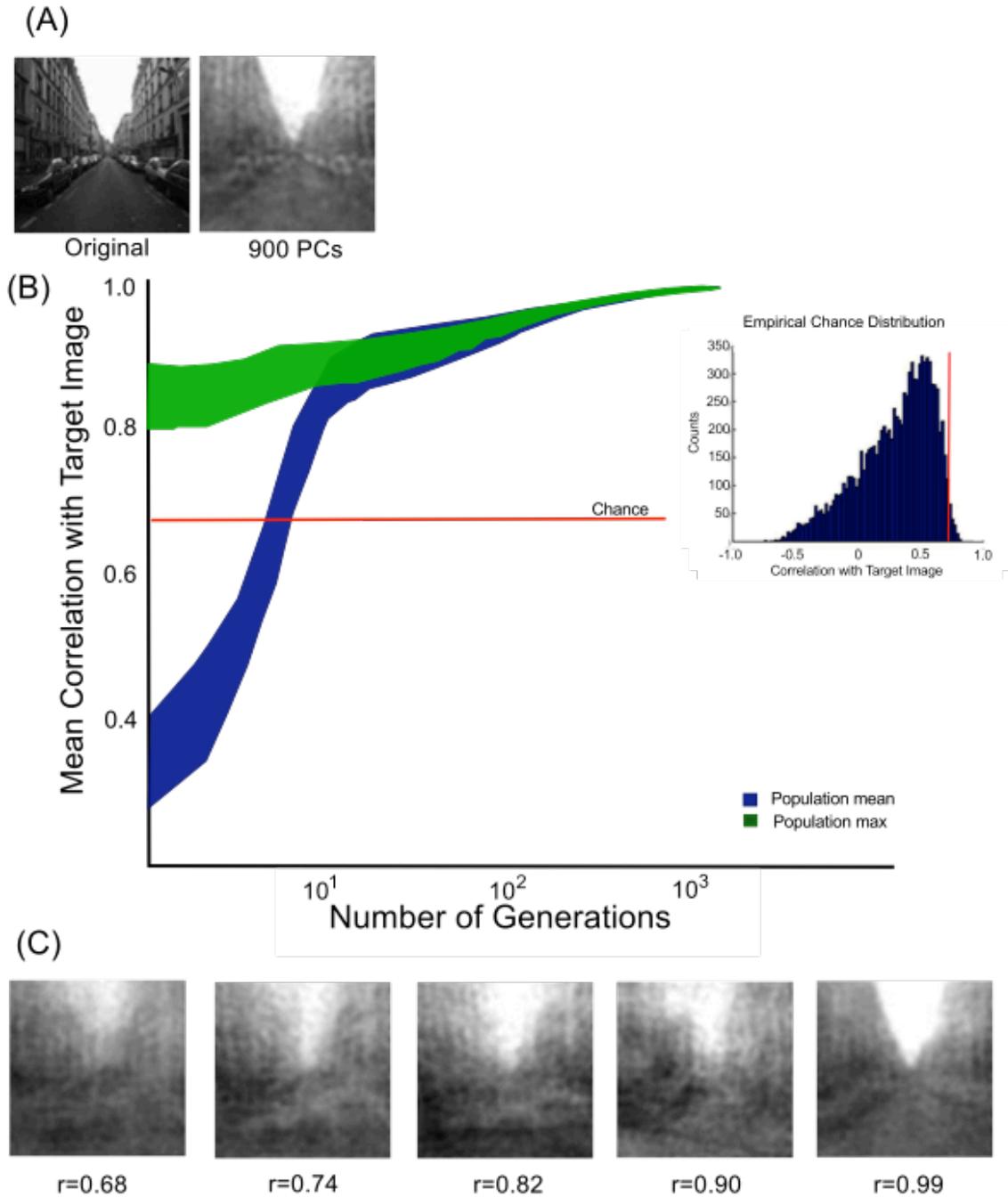

**Figure 2: Results from ideal observer simulations. (A)** Original image reconstructed by ideal observer (left) and the same image represented with its first 900 principal components (right). The right image represents the upper limit of reconstruction fidelity. **(B)** Population mean (blue) and max (green), including 95% confidence intervals, for the ideal observer. The red red line represents chance-level performance from the empirical chance distribution (inset). This distribution represents the correlations of 10,000 random noise images with the target image. The red line is drawn at the 95th

percentile (r=0.68). (C) Intermediate reconstructions representing various levels of correlation with target image.

### *Human-Computer Collaborative Image Reconstruction*

As each generation in the simulation involved approximately 250 pairwise comparisons, a typical run of 1,059 generations results in over 250,000 comparisons. It is therefore not practical for a single human observer to reconstruct an image to the r=0.99 level. However, we observed that by the fifth simulated generation, the best images of the population were highly correlated with the target image (see Figure 2B). Our next objective therefore was to determine how well human observers could re-create a specific photograph in five generations (1250 comparisons, or about an hour's worth of work).

Figure 3 shows three reconstructed photographs from five separate human observers. The three images that were not a part of the PCA database, and included the street scene used for the ideal observer. Participants were able to accurately re-create these photographs, even in this small amount of time. For each image, at least two observers achieved a statistically significant level of reconstruction (an image more similar to the target than 95% of random noise images). When considering only the street scene and mountain scene (see Figure 3), four of five observers achieved this level of reconstruction. Averaging the reconstructed images across observers resulted in images that were more similar than 99% of randomly drawn images, and better than the best single observers, suggesting that each individual observer succeeded in reconstructing slightly different image information.

Can we predict what photograph an observer was emulating from the resulting reconstructed image? A correlation classifier revealed that we could predict what image each participant was reconstructing at a level that was highly above chance (80% correct, chance=33%, p<0.05, binomial test). Therefore, even with a relatively small number of trials, it is possible for human observers to reconstruct a specific image using our method.

We examined the trial-by-trial performance of each human observer to determine how optimally each observer classified the noise images. For each trial, we correlated each noise image to the target, and determined whether the observer had chosen the best image. On average, human observers chose the best image on 79% of trials (range: 66%-86%). We sorted trials by the similarity of image pairs, and then binned them into quintiles. We found that although human performance increased with decreasing pair similarity, observers remained above chance even for the most similar bin (66% correct, t(14)=68, p<0.0001, see Table 1). In other words, even in the most challenging cases human observers were able to not only see target information in the noise but also choose optimally among very similar competitors.

| Bin | 20th | 40th | 60th | 80th | 99th |
| --- | --- | --- | --- | --- | --- |
| **Mean** | 0.66 | 0.70 | 0.74 | 0.82 | 1 |
| **Stdev** | 0.04 | 0.06 | 0.09 | 0.15 | 0 |

**Table 1: Proportion of trials in which human observers adopt ideal choice behavior as a function of image pair similarity. Bins are ordered from most similar to least similar.**

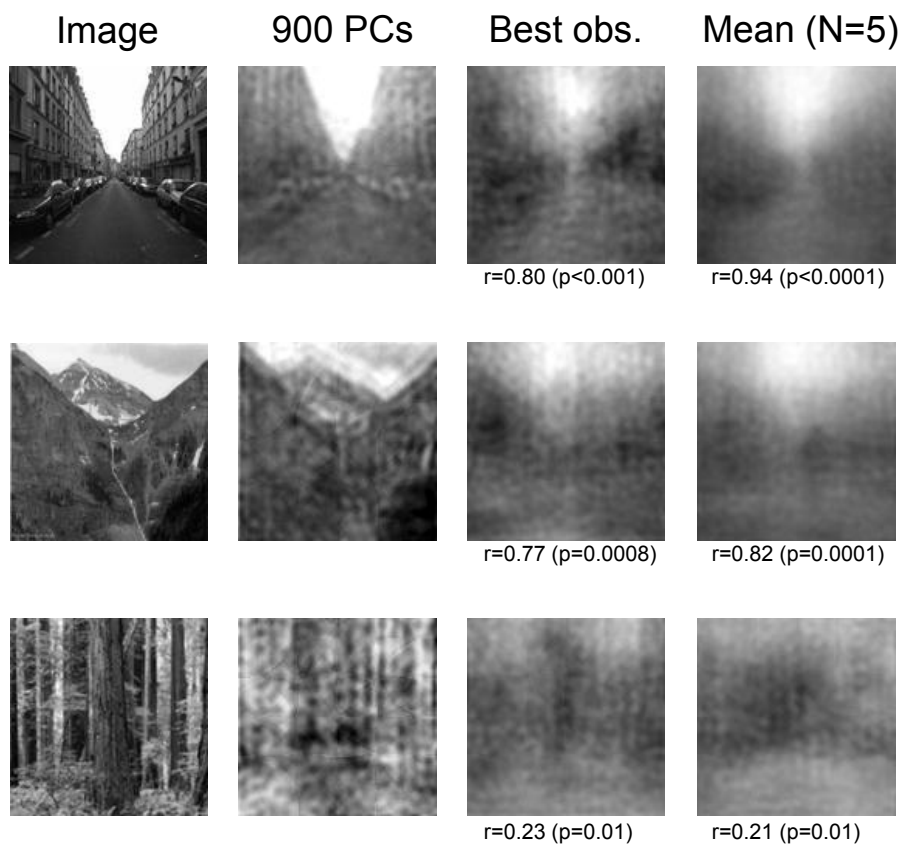

**Figure 3:** Five human observers reconstructed one of the three scenes in the left-hand column. The quality of reconstruction is bounded by the representation of the image in PC space (second column). The third column shows the reconstructed image of the best observer, and the last column shows the averaged reconstruction for all five observers. P-values represent probability that a randomly created noise image is more similar to the target under the empirical chance distributions.

Although observers had success in reconstructing each of the images relative to the empirical chance distributions, we obtained numerically higher correlations with street scenes. To what extent is this due to a bias in the statistics of the visual noise? To address this issue, further ideal observer analyses revealed that our method could be used to fully reconstruct arbitrary scenes, even indoor images with extremely different statistical properties (see Supplemental Materials).

### *REVEALing observers' mental image of a street scene*

Can REVEAL be used to visualize a purely mental image of a scene? Five participants were instructed to think of a typical street scene from the point of view of a driver driving down the road. As we did not know in advance how long it would take for a good image to emerge, participants were allowed to terminate the experiment when they were satisfied with the result. Observers chose to terminate the experiment after 853-1801 trials (3.5-7.2 generations).

The reconstructed street images for each participant are shown in Figure 4. Although each image contains fewer details than a photograph, we can clearly see street-like spatial layouts in each of the images. We quantitatively evaluated the reconstructions results by performing a nearest neighbor image retrieval, using three databases: (1) the 4200-image database used for PCA; (2) the entire SUN database, a comprehensive scene database of over 900 scene categories (16); and (3) a database of over 20,000 street scenes in order to better compare the reconstructions to real-world scenes. The most similar images for each database are shown next to each reconstruction in Figure 4. For four of five observers, the 20 most similar images from the PCA database were almost exclusively streets or rural roads (mean: 90% streets, range: 70%-100%), while one observer's most similar images were all mountain scenes (Observer 5 in Figure 4). Even among the 900 categories of the full SUN database, the 20 most similar images were either streets or rural roads 52% of cases. This is in stark contrast to the 10% street retrievals that could be expected by chance. (Chance level here was determined by taking 10,000 bootstrap samples in which 20 images were randomly selected from the >129,000 image SUN database). These simulations suggest that the reconstructed images for each observer were much more similar to actual real-world street scenes than what can be expected from chance alone.

We now visualize the type of street scene envisioned by each observer by comparing each observer's reconstruction to 20,000 real-world street scenes. This also aids in evaluating the quality of the reconstructions. For example, Observer 5 reconstructed an image that was not similar to any of the street scenes in the PCA database, but was highly correlated with other types of street scenes (see Figure 4). In order to compare these observed similarities with what might be expected from chance, we created 10,000 noise images and correlated each of them to each of the 20,000 street scene images, and observed the maximum correlation in the database. The distribution of maximum correlation values reflects the correlations that could be expected to *any* street scene from *any* noise image, and is therefore a very stringent chance criterion. Despite this severity, we found that three of the five observers had retrieval correlations better than 99% of the randomly drawn images. Observer 2 had a retrieval correlation better than 70% of randomly drawn images, while Observer 4 had a retrieval correlation only better than 3% of randomly drawn images.

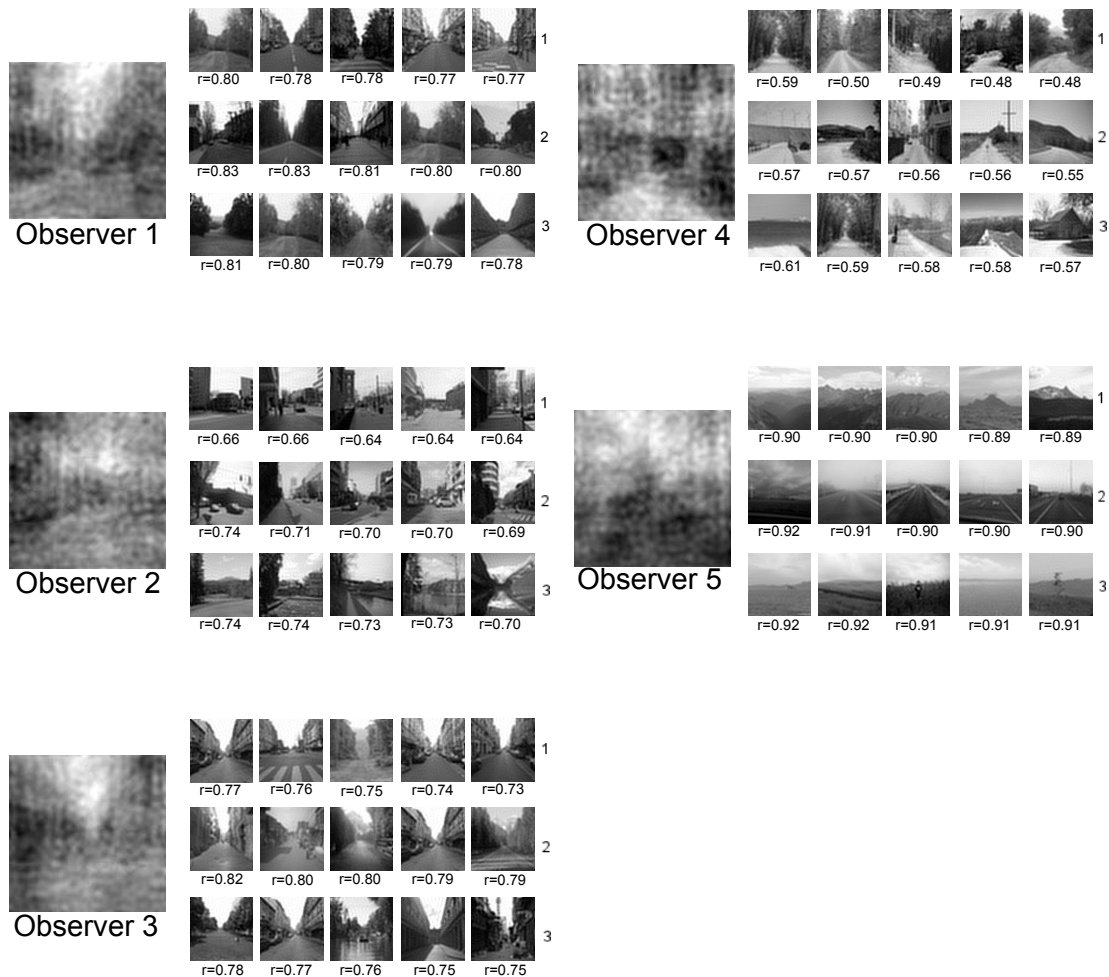

Figure 4: Reconstructed street scenes from each of the five observers. For each observer, we show the five most similar images from the PCA database (top), the 20,000-street scene database (middle) and the SUN database (bottom row).

To what extent does one's ability to create vivid mental images influence the quality of one's reconstruction? Each of the five observers also completed a standard vividness of visual imagery questionnaire (VVIQ, (17, 18)) following the reconstruction experiment. For simplicity, we have inverted the original scale such that low VVIQ reflects low imagery abilities. Our five observers ranged from 2.2 to 3.5 on the inverted 1-5 VVIQ scale, corresponding to slightly below average to very good mental imagery abilities (19). The correlation between the inverted VVIQ and the quality of the reconstruction (measured as the maximum correlation between reconstructed image and most similar image in the 20,000 street database) was r=0.69. Therefore, one's ability to create a clear mental image might set a limit on the quality of the reconstructed images we can make.

To what extent do the reconstructed street scenes reflect representations that are unique to the observer? On the one hand, one's mental image of a street could be a particular street the one has experienced, such as the street one grew up on. On the other hand, observers might have shared street representations, perhaps reflecting the central tendency of all of the streets one has experienced. We correlated each observer's reconstructed image to the reconstructed images of the other four observers. We found that the mean correlation between each pair of observers was r=0.38 (range: 0.14 to 0.66, see Supplementary Materials for the full correlation matrix between observers). This moderate degree of resemblance indicates that each observer's representation has idiosyncratic elements.

An alternative possibility is that observers share similar street representations, but have different degrees of success in reconstructing them. In particular, we correlated each participant's reconstructed street with the averaged image of the 20,000-street database. Prototype models of categorization assert that our shared category representations reflect the central tendency of a category, so the averaged image can be considered a proxy for this central tendency. We found a sizeable resemblance between individual street representations and the central tendency of the category (mean: r=0.58, range: 0.30-0.81). Interestingly, however, the extent to which an individual observer's reconstruction matched the average image was highly correlated with that observer's inverted VVIQ (r=0.93, see Figure 5), suggesting that although observers may have had similar street representations in mind they differed in their ability to reconstruct them. This variability was substantial - the slope of the regression line between the observers match to the central tendency and their VVIQ was 0.4, indicating that every unit increase in VVIQ resulted in an increased correlation with the average street image of 0.4

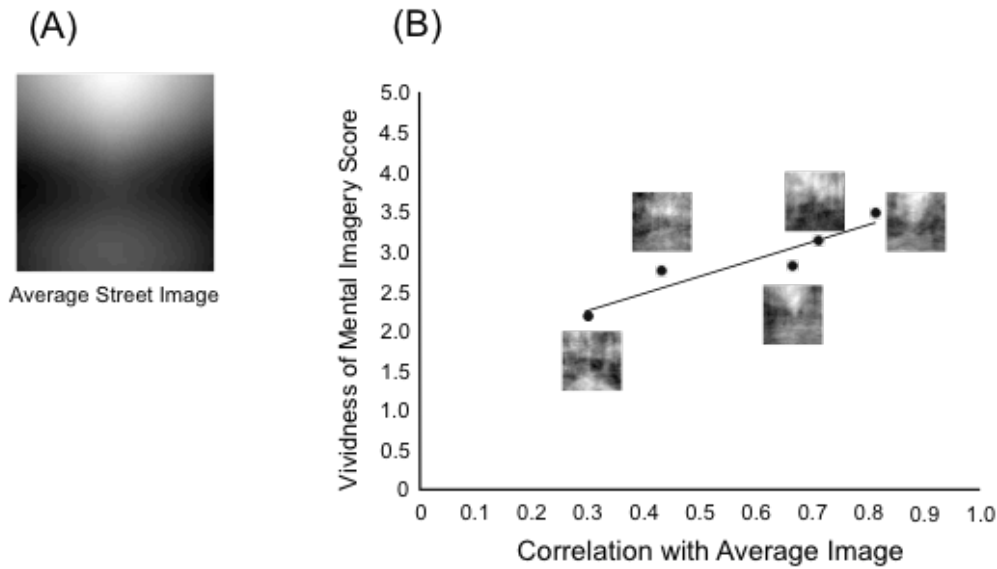

**Figure 5: Visual imagery vividness scores (VVIQ) predict the similarity of an observer's mental representation to the average image. (A) The average image of the 20,000-street database. (B) The resemblance of each observer's reconstructed street scene with the average image in (A) closely follows the observer's vividness of mental imagery score. Note that we have inverted the original vividness of visual imagery (VVIQ) scale such that a high degree of mental imagery receives a larger score.**

### *Predicting rapid scene understanding from reconstructed images*

To what extent do the reconstructed street scenes reflect a category template that an observer can use for rapid scene recognition? If a reconstructed image reflects an observer's mental template, we would expect more efficient perception of real-world images that resemble this template. We tested this prediction directly in a short detection experiment that took place immediately after the reconstruction experiment.

Participants briefly viewed either street scene images or fully phase-randomized versions of these images, and indicated whether each image was an intact street scene or phase-scrambled image. Half of the street scenes were images that were the most similar to that participant's reconstruction while half were the least similar street scenes. The presentation times for each participant were chosen from a preliminary block aimed at determining the participant's 75% detection threshold for presentation time.

We computed detection sensitivity (d') for both the most similar and least similar images separately and found that observers were more sensitive in detecting street images that were more similar to their reconstructed image (mean d'=4.25) than images that were the least similar to their reconstructed image (mean d'=2.11, $t(3)=5.8$, $p<0.05$, see Figure 6b). One participant (Observer 5) was omitted from the accuracy analysis due to ceiling performance resulting from an overestimated presentation time threshold (see Supplementary Materials for details). Therefore, images that are more similar to an observer's reconstructed image are also detected better during rapid visual presentations, suggesting that our reconstructed images not only resemble observers' internal scene category representations but that observers also use those representations to aid in perception.

Although observers were not specifically instructed to minimize response times, we examined reaction times so that we could include Observer 5 whose accuracy was at ceiling. We found that each observer was faster to categorize the street scenes that were most similar to his reconstruction, compared to the least, and that on average, reaction times to the most similar images were 8.5% faster, see Figure 6c).

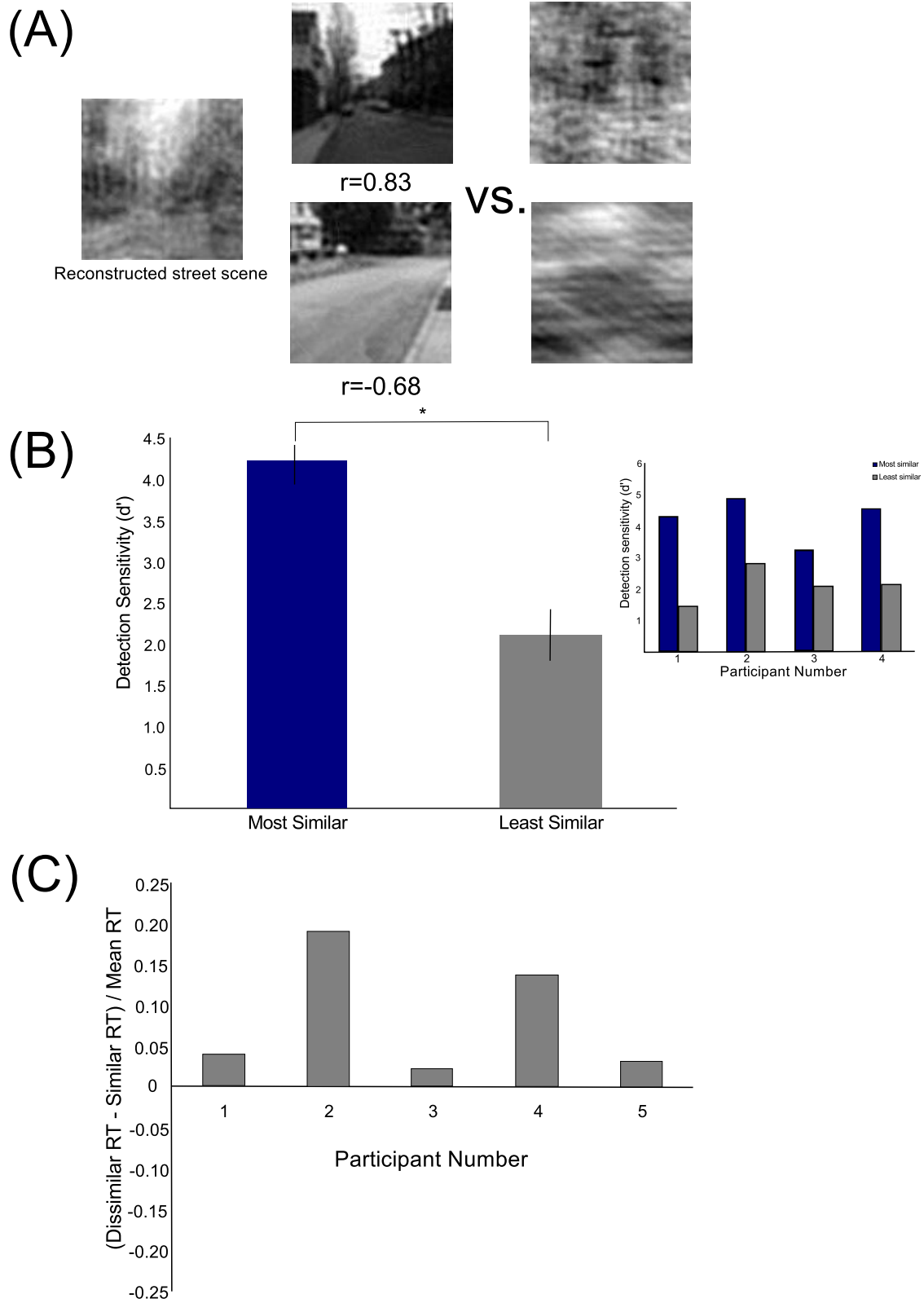

**Figure 6: Rapid detection experimental design and results.** Observer 1's

**reconstruction is used for the purposes of illustration. (A) Example stimuli for rapid detection selected from 20 street scenes that were the most similar to an observer's reconstruction (top) and the 20 street scenes that were least similar (bottom) Observers classified images as intact scenes (left) or phase-randomized images (right). (B) Detection sensitivity (d') for the most and least similar images. Inset shows detection data from individual observers. Observer 5 was omitted due to an error in threshold estimation. (C) Least similar reaction times minus most similar reaction times, normalized by the average RT of each observers. All observers were fastest to classify the images that were the most similar to their own reconstructions.**

Rapid scene categorization performance can be predicted by information in an observer's reconstructed mental image. Observers were faster and more accurate in detecting images that were more similar to their reconstructed template, suggesting that these reconstructed images might serve as a high-level visual template that can be used to rapidly classify visual scenes.

**Discussion**

A hallmark of human cognition is our ability to form vivid mental images in the absence of visual input. In this work, we have introduced REVEAL, a method that allows us to visualize these mental images for the first time. Visualizing this information not only allows us to examine the subtle, often ineffable visual cues that make up scene identity, but also allows us to gain insights into the mechanisms of rapid scene understanding. By presenting observers with random visual inputs, we are able to obtain their internal scene representations in an unbiased manner.

Although previous work has been able to visualize internal representations of simple stimuli using reverse correlation (12, 20), several innovations were needed to make this possible for complex, real-world scenes. Techniques such as classification images have been used to successfully uncover mental representations of simple stimuli. However, they have several drawbacks that make them impossible to use for real-world scenes (21): first, classification images cannot deal with correlations in noise patterns, or between noise patterns and stimuli, as is the case with natural scenes (22, 23). Second, even an image of moderate resolution will have tens of thousands of pixels, requiring possibly hundreds of thousands to trials to adequately sample each. Finally, the Gaussian white noise used in classification images does not have a very natural appearance, and this may change this visual strategies used by observers to perform the classification (24). Similarly, it is not clear that the information used by observers would correspond to pixels at all, as attention seems to follow surfaces (25) and objects (26). Although classification images provide templates of decision boundaries used by observers, they cannot reveal a category's central tendency (14), nor can they make predictions about the perception of subsequent stimuli (13, 15). While more modern approaches, such as Bubbles (24, 27) ameliorate some of these issues, they still seem to perform best on stimuli for which the relevant features can be intuited, such as the emotional expression of a face. Similarly, Olman & Kersten (13) present the idea that reverse correlation can be performed on any set of parameterized stimuli, and term this approach "classification objects". However, it has been difficult to create a generative model for something as complex as a real-world scene. Other generative approaches that sample over category probability distributions may not converge for real-world categories with many dimensions (14). By contrast, our method uses collaboration between human observers and a machine learning algorithm in order to efficiently traverse the space of natural images, creating an image that reflects the central tendency of the category with a relatively small number of trials. Similar genetic algorithms have been used successfully in physiology to map out response properties of neurons in V4 and IT in a principled and efficient manner (28, 29).

Although human observers have exceptional performance in scene understanding, strong prototype effects are also observed for scenes. Interestingly, images that have been rated as better exemplars of their categories are recognized more efficiently by human observers than less typical instances (30, 31); are classified with higher accuracy by machine vision systems (32); and evoke more consistent patterns of brain activity that can be more easily decoded by a classifier (31). This work not only confirms these observations, but also extends them by visualizing and predicting the images that will show facilitated classification.

Several aspects of this study hint at a prototype structure for scene categories. The quality of an observer's reconstructed image was predicted by individual differences in visual imagery abilities, and observers with strong visual imagery abilities reconstructed images that were more similar to the central tendency of the category. On the other hand, observers' reconstructed images only bore moderate resemblance to one another, leaving open the possibility that one's own representation contains information distinctive to an individual. Therefore, future work will investigate the extent to which scene categories may be represented with prototypes. Comparing current visual input to a stored prototype is an efficient way to code complex visual signals as fewer resources are spent coding the things that are most frequently experienced (33, 34). There is evidence that some stimuli, such as faces, are explicitly coded in reference to the average, or prototype (35–37). Typically, prototype referencing is revealed through adaptation. As adaptation to visual scenes can produce robust aftereffects (38, 39), this hypothesis can be directly tested.

In conclusion, we developed REVEAL, a novel method for visualizing observers' subjective mental representations of real-world environments. This provides us with the rare opportunity to get a look at the private, mental representations of another person. In addition to providing compelling visual images, we have shown that the content of these images can predict the rapid scene detection performance of an observer on an image-by-image basis. By visualizing complex mental representations, we leave the door open for future work to examine the role of these representations in rapid perception, attention and memory. More broadly, externalizing an internal representation fulfills a key goal of cognitive science: understanding how neural representations are formed and the nature of the information used to form them.

**Methods**

### *Creating visual noise from natural scene statistics*

In order to create visual noise based in natural image statistics, we amassed a database of 4200 natural scene images of street, forest, and mountain environments, taken from the SUN database (16). We represented each image in this database as the output of a bank of multi-scale Gabor filters. This type of representation has been used to successfully model the representation in early visual areas (40). Each image was converted to grayscale, down sampled to 128 by 128 pixels, and represented with a bank of Gabor filters at three spatial scales (3, 6 and 11 cycles per image with a luminance-only wavelet that covers the entire image), four orientations (0, 45, 90 and 135 degrees) and two quadrature phases (0 and 90 degrees). An isotropic Gaussian mask was used for each wavelet, with its size relative to spatial frequency such that each wavelet has a spatial frequency bandwidth of 1 octave and an orientation bandwidth of 41 degrees. Wavelets were truncated to lie within the borders of the image. Thus, each image is represented by 3*3*2*4+6*6*2*4+11*11*2*4 = 1328 total Gabor wavelets. The weight of each Gabor for each image was determined using ridge regression.

Although the Gabor pyramid reduced the dimensionality of the representation from 16,384 pixels to 1,328 Gabors, we wanted to achieve a higher level of dimensionality reduction. We performed principal components analysis (PCA) on the 4200-image by 1328-wavelet weight matrix. A pilot experiment conducted on Amazon Mechanical Turk determined that images could be represented with the first 900 principal components without loss of categorization accuracy (see Supplementary materials for details).

Noise images were created by choosing random values for each principal component score, scaled to the observed range for each component.

***Establishing chance-level performance empirically***

In order to evaluate our success in image reconstruction, we needed to know what level of reconstruction fidelity could be expected from chance alone. We determined this chance level empirically by creating 10,000 random noise images and then evaluating their pixel-wise correlations with each of the three target images in the first experiment (see Figures 2-3). If an experimentally observed correlation was higher than 95% of all random correlations, then the reconstruction was considered significantly better than chance. The three chance distributions were substantially dissimilar from one another, and the Supplemental Materials contain a characterization of these differences and a discussion of the possible implications of this fact.

### *Genetic Algorithm*

A flowchart of the genetic algorithm can be found in Figure 1C. We began by creating an initial population of 100 noise images. For the ideal observer and human observers in the first experiment, we wanted to ensure that any successful reconstruction could not be due to a particularly auspicious initial population. Therefore, we rejected any initial images that were above the 80th percentile of similarity from the empirical chance distribution (see above). As we could not know in advance what visual features observers would use in the second experiment, the initial population of images was randomly generated.

Human observers viewed pairs of noise images and performed a 2AFC task indicating which image was a better match to the target. As evaluating every pairwise comparison of each 100-image generation would require nearly 5,000 trials per generation, pairs were sampled such that each of the 100 images was viewed five times in a generation.

At the end of a generation (about 250 trials), individual images were selected to advance to the next generation using a roulette wheel mating strategy (41). Specifically, each image from the previous generation was replicated according to the number of times it was selected, and 100 new images were randomly sampled to become a proto-generation from this larger pool. Therefore, images that were frequently chosen by observers had a better chance at advancing to the next generation, while images that were chosen infrequently advanced infrequently. In order to introduce more variability into the new population, a "cross-over" step was performed with 0.4 probability. An image selected for cross-over was combined with another randomly selected image from the proto-population. The resulting "child" image had the odd principal components of one "parent" image and the even principal components of the other. Additionally, images were "mutated" with 0.3 probability. During mutation, random values were added to the principal component scores of the image, walking it a small step away from the original image in feature space. The random values were scaled to reflect 5% of the standard deviation in PC values. Last, new randomly generated noise images replaced a portion of the new generation's images. For the second generation, this probability was 0.6, and was halved in each subsequent generation. The probabilities for crossover, mutation and replacement were chosen on the basis of computer simulations and pilot experiments.

The first experiment terminated after five generations. Participants in the second experiment were allowed to terminate the experiment seeing a satisfactory result. This terminal image was then compared to a database of 20,092 street scenes using pixel-wise correlation, and the 20 street scenes most similar to the reconstruction and the 20 scenes least similar to the reconstruction were saved for use evaluating the reconstructions using the rapid detection task.

### *Experimental methods: psychophysics*

**Participants:** Eleven individuals (6 female, ages 20-31), including the authors MRG and AB took part in the first experiment. Each author participant took part in reconstructing each of the three target images while the nine remaining naïve participants took part in only one of the three.

In the second experiment, six participants (1 female, ages 23-31) took part, including the authors MRG and AB. One of the four non-author participants was omitted from analysis for abusing the self-termination system (see Supplementary Materials). Therefore, the remaining participants were chosen to be trusted, experienced psychophysical observers from the Stanford University research community.

**Procedure:** Participants were seated in a dimly lit room approximately 54 cm away from a 21-inch CRT monitor (Sony Trinitron, Tokyo Japan). Participants were told that they would see pairs of images, and to indicate with a keypress which of the images was more similar to the target. For participants in the first experiment, the target image was shown above the two noise images, and for participants in the second experiment, the word "street" was displayed instead. Participants in this experiment were told to envision a prototypical street scene as if they were in the scene driving down the street. Images were square and subtended approximately 5 degrees of visual angle on each side. There was approximately 1 degree of visual angle between the two images. Images remained on the screen until the participants responded. Participants were made aware that they would be collaborating with a computer algorithm that was trying to learn their visual strategy, and to expect images to get better over time. Participants were not put under time pressure to respond quickly, and reaction times were not recorded. At the end of each generation (~250 trials), participants were allowed to take a break.

**Rapid detection experiment:** Between the reconstruction experiment and the detection experiment, participants were given a ten-minute break. During this time, their final reconstructed images were compared to each of the images in the 20,092-street scene database, and 20 of the most similar images and 20 of the least similar images were selected as targets in the detection experiment. In addition, 50 random images were chosen for use in measuring each participant's presentation duration threshold.

Following the break, participants were told that they would do a quick detection experiment in which they would see rapidly presented images, and that they should classify the images as either intact street scenes or meaningless visual noise using a keypress. The goal of the first block of 100 trials was to determine the presentation time necessary for the observer to achieve 75% correct on the random images. Participants viewed the 50 randomly selected street scenes and 50 phase-randomized versions of these scenes in random order. Each trial commenced with a fixation point for 200 msec, followed by the experimental image and then a dynamic pattern mask (42). The initial presentation time was set to 50 msec. Three sequential correct responses resulted in a decrease of subsequent presentation times by 10 msec (to a floor of 10 msec), while incorrect responses resulted in an increased presentation time by the same amount (to a ceiling of 200 msec). This procedure will converge on the presentation time necessary to achieve 75% correct performance. At the end of the 100 trials, we defined the presentation time threshold for each observer to be the second-lowest presentation time viewed in the block.

After the presentation time threshold was determined, each participant completed an experimental block of 80 trials using the same procedure outlined above except that all presentation times were set to the threshold obtained in the previous trial. Of the experimental images, half of the street scenes were the 20 images most similar to the participant's reconstruction while the other half consisted of the 20 images that were least similar to the reconstruction. As images in both groups of scenes were street scenes, any difference in detection accuracy must stem from the perceptual distance of the image to the participant's mental image of the category. Participants were instructed to respond as quickly and accurately as possible. Feedback on performance was not given.

**Supplemental Materials**

*1. Determining the number of principal components to use*

In order to determine the number of principal components we would use for this experiment, we performed the following categorization study on Amazon's Mechanical Turk (AMT). The goal of the experiment was to determine the minimal number of principal components necessary to represent the images without loss of categorization accuracy. We used 99 images of street, mountain and forest environments (33 images each). None of these images were from the image database used for PCA. Each of the 99 images were represented with their first 64, 128, 256, 500, 600, 700, 800, 900, 1000 or 1328 principal components. Each image was presented to two observers on Amazon Mechanical Turk, who classified the images into one of the three categories (forest, mountain street). As shown in Figure S1, classification performance was at chance for images represented with fewer than 500 PCs, and as expected, classification performance was at ceiling when no principal components were removed. We chose to move forward with 900 principal components as this was the smallest number of PCs that could be used without loss of classification accuracy.

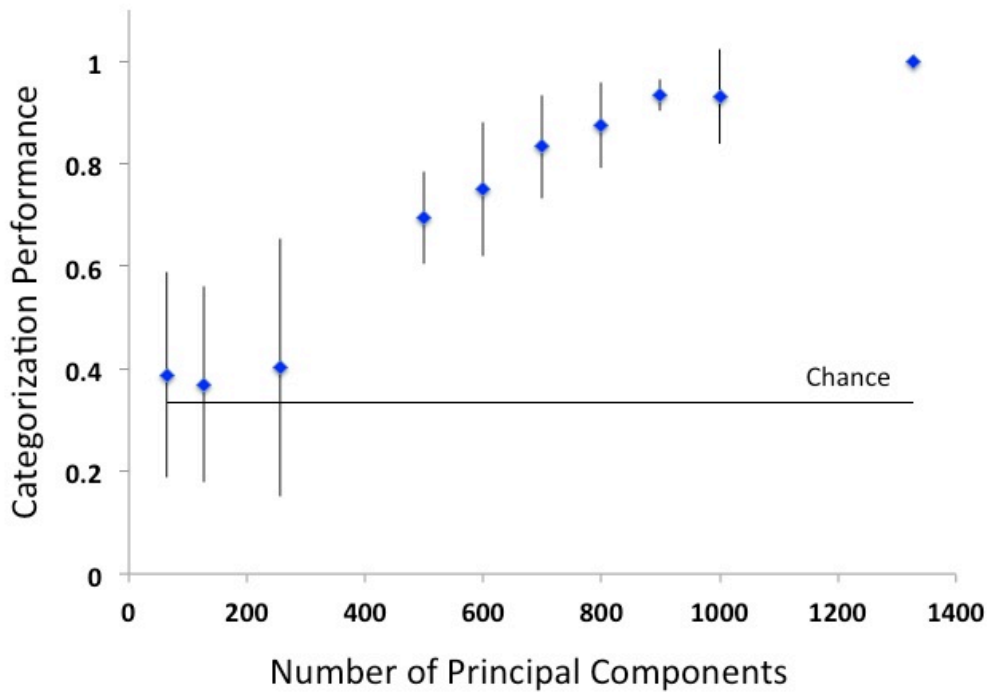

**Figure S1: Results from online pilot study aimed at determining the number of principal components necessary to categorize a scene. We chose 900 components as this was the least number of components without loss of categorization accuracy.**

## *2. Comparison of our method with "superstitious" classification images*

Our experimental method represents a departure from what has been done in "superstitious" classification images in terms of both noise type and experimental paradigm. In order to determine whether we could obtain similar results using a classic classification images paradigm, we simulated two observers in a task similar to that of Gosselin & Schyns (2003).

Each simulated observer viewed 20,000 trials in which a single noise image was evaluated in terms of similarity with the street scene in Figure 2A. The simulated observers either accepted or rejected each trial based on how similar the noise image was to the target scene (using pixel-wise correlation as the similarity metric). One simulated observer followed a strict acceptance criterion and only accepted images that were better than 90% of images from the empirical chance distribution. The other simulated observer accepted images that were better than 60% of images from the chance distribution. These acceptance rates were chosen to be in the range of human observers in Gosselin & Schyns (2003). For each simulated observer, a classification image was created using a linear model.

The strict observer accepted 4,654 of the 20,000 trials, and the resulting classification image was correlated r=0.90 with the target image. The lenient observer accepted 9,764 of the 20,000 trials, and this resulting classification image was correlated r=0.89 with the target image. Both of these correlations are better than any observed value from the empirical chance distribution, indicating that this method can successfully reconstruct this image. However, when we compare these simulations to the genetic algorithm simulation, we find that the genetic algorithm could reach this level of performance in 37 generations * 250 trials = 9,250 total trials. Therefore, similar performance can be reached in less than half the trials by using our method.

*3. Comparison to genetic algorithm using Gaussian white noise*

The analysis outlined above demonstrates that superstitious classification images can yield a reasonable result with our noise that is derived from natural scene statistics. How much does this noise contribute to our overall result? To address this question, we simulated an ideal observer that was identical to the ideal observer presented in the main text, except that the observer viewed white noise images. We ran the simulation for ~1.3 million generations, and achieved a maximum correlation of r=0.85. As performance increased linearly after 1 million trials, we extrapolate that we could achieve r=0.99 level correlation in 2.8 million generations. As a generation involves ~250 classifications, this is over 702 million trials, and thus far out of reach of any human experiment. This result emphasizes the utility of our visual noise for reducing the effective dimensionality of the problem.

*4. Rejection of human participants*

One observer was omitted from the second experiment due to abusing the self-termination system. This participant quit after 460 trials (mean of other subjects: 1337 trials, range: 853-1801). The mean correlation of this participant's final image with the 20,000 street scene database was -0.05, suggesting that the participant did not construct anything very street-like. This correlation was significantly lower than that of the remaining participants ($r=0.20$, $t(4)=8.1$, $p<0.005$).

One observer (Observer 5) was omitted from the detection accuracy analysis of the rapid detection experiment for near-ceiling performance (93%) that was significantly higher from that of the remaining participants who were between 75% and 80% correct. We estimated the presentation time necessary to achieve 75% correct in this task, and set the experimental presentation time to this value. We believe that we overestimated this participant's threshold. We included this participant in the reaction time analysis, however.

### *5. Different chance distributions for different images*

Although equally high reconstruction correlations could not be achieved for all three of the images that we used, participants were still achieving similar performance under the empirical chance distributions, see Figure S2 and also Figure 3. Although none of these three images were in the image database used to create the noise, the street scene in Figures 2 and 3 turned out to be more similar to images in the PCA database than the other two images. Furthermore, as many of the forest images in the PCA database had trails or paths, the average image of the PCA database was very street-like, explaining the better success in recreating this image compared to the others.

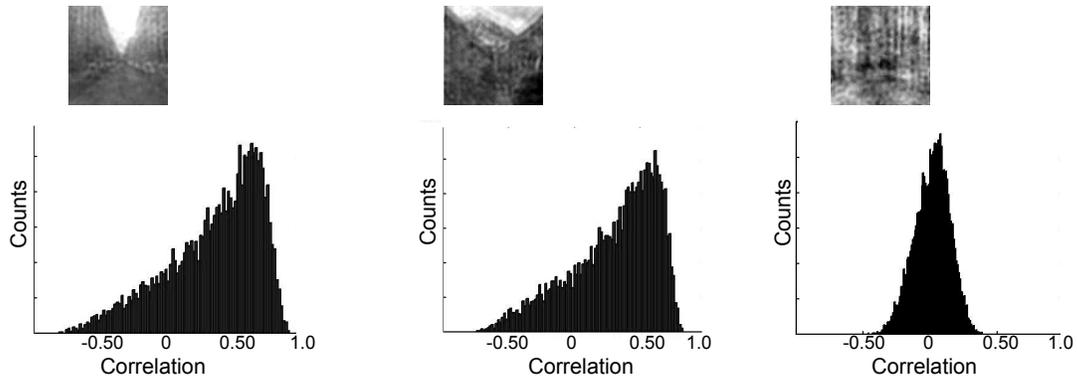

**Figure S2: Empirical chance distributions for each of the three images used in the first experiment.**

In order to show that our method does not strongly depend on the particular images making up the PCA database, we ran the ideal observer on the least correlated image, the Redwood forest image shown on the far right of Figure S2 (see also Figure 3). The ideal observer was able to construct an image correlated with the original at the r=0.99 level in 31,907 generations. Although this is substantially less efficient than the 1,059 generations needed for the street image on the left-hand side of Figure S2, this simulation demonstrates that it is still possible to reconstruct an image whose features are not highly correlated with images in the PCA database. To further probe this point, we ran a third ideal observer that tried to reconstruct an image of a living room. As the PCA database consisted of images of streets, mountains and forests, this image represented a further departure from the statistics of the PCA database. We found that the ideal observer could reconstruct a scene more similar to the input than 99.999% of randomly drawn scenes in 25 generations, and could reach the r=0.91 level of reconstruction in 77,061 generations. Extrapolating from this curve suggests that the r=0.99 level could be reached in approximately 1.2 million generations. Therefore, although the PCA database does influence how easily an image can be reconstructed, our method can be used to reconstruct a broad array of images. This fact can be harnessed in future experiments by creating a PCA database reflecting the statistics of what one wants to reconstruct.